\documentclass[num-refs]{wiley-article}

\usepackage{siunitx}

\papertype{Original Article}
\paperfield{Journal Section}

\usepackage{natbib}
\usepackage{booktabs}
\usepackage{todonotes}
\usepackage{subfig}

\newcommand{\round}[1]{\ensuremath{\lfloor#1\rceil}}
\newcommand{\etal}{\textit{et al}. }

\title{Feature embedding in click-through rate prediction}

\author[1]{Samo Pahor}
\author[1]{Davorin Kopič}
\author[2]{Jure Demšar}

\affil[1]{Zemanta, an Outbrain company, 1000 Ljubljana, Slovenia}
\affil[2]{Faculty of Computer and Information Science, University of Ljubljana, 1000 Ljubljana, Slovenia}

\corraddress{Jure Demšar, Faculty of Computer and Information Science, University of Ljubljana, 1000 Ljubljana, Slovenia}
\corremail{jure.demsar@fri.uni-lj.si}

\fundinginfo{The research at hand did not receive any institutional funding.}

\runningauthor{Pahor et al.}

\begin{document}

\begin{frontmatter}
\maketitle

\begin{abstract}

We tackle the challenge of feature embedding for the purposes of improving the click-through rate prediction process. We select three models: logistic regression, factorization machines and deep factorization machines, as our baselines and propose five different feature embedding modules: embedding scaling, FM embedding, embedding encoding, NN embedding and the embedding reweighting module. The embedding modules act as a way to improve baseline model feature embeddings and are trained alongside the rest of the model parameters in an end-to-end manner. Each module is individually added to a baseline model to obtain a new augmented model. We test the predictive performance of our augmented models on a publicly accessible dataset used for benchmarking click-through rate prediction models. Our results show that several proposed embedding modules provide an important increase in predictive performance without a drastic increase in training time. 

\end{abstract}
\end{frontmatter}

\section{introduction}

Online advertising, as opposed to traditional advertising (e.g., in newspapers, on billboards, on television, on radio, etc.) is typically featured on websites or mobile applications. Online advertising allows companies to reach a worldwide user base and engage key demographics to market their products. It offers a way for companies to increase brand awareness as well as their understanding of target audiences. For online advertising to succeed, ads need to be prominently featured and presented to relevant consumers. The vast majority of online advertising space is sold through programmatic advertising. In programmatic advertising, whenever a user opens a web page or an app, an auction is executed in the background for each ad space on that page. The ad space in these real-time bidding (RTB) auctions is thus being dynamically sold to the highest bidder \cite{edelman2007internet,varian2007position}. Bidders are typically specialized companies that offer their services to advertisers in order to participate in RTB auction as efficiently as possible. Since loading of web pages needs to be as fast as possible and any delays would ruin the user's browsing experience, a key characteristic of RTB auctions is that they are close to instantaneous, occurring in less than 100 milliseconds \cite{6960761}. The advertisement space in programmatic advertising is completely automatized, the whole advertising process is being performed, managed and optimized by software directly.

The ad that wins the auction is displayed to the user visiting the website, in the hopes that said user will respond positively \cite{Singh2009BlockingOA} and potentially click on the ad. Since such a click directly translates to online traffic and therefore value for the advertiser, predicting click-through rate (CTR) is a key part of the advertising businesses \cite{richardson2007predicting}.  CTR prediction demands very quick response rates, so it is able to function within the RTB environment. Additionally, the amount of constantly generated information in the online advertising space is overwhelming and impossible to organize or oversee manually. Both issues, quick response rate and overwhelming amount of data, are addressed with sophisticated machine learning models, which are able to extract important knowledge from past data, are trained on-the-fly and able to perform extremely fast predictions.

Models predict user clicks based on a variety of inputs, including contextual data related to the user, historical data and other data which varies across domains. Mentioned input features are predominantly categorical and can usually take on a very large amount of different values. A typical approach with such data is to transform them into high-dimension sparse binary vectors via one-hot encoding. These sparse vectors can be problematic for some of the more popular and modern machine learning techniques, such as deep learning. In these cases further embedding and dimensionality reduction is necessary to produce dense numeric feature vectors that can then be used in model training. 

Recent research in the field of CTR focuses on optimizing prediction models, such as logistic regression (LR) \cite{41159}, factorization machines (FM) \cite{rendlefm}, deep neural networks (DNN) \cite{DNNarticle} or hybrid approaches, like DeepFM \cite{guo2017deepfm}. LR is essentially a linear model, assigning a specific weight to each observed feature and using a logistic function for final prediction. A more sophisticated method are FMs, where the linear model is upgraded with an additional interaction term. This gives FMs the ability to capture 2nd order interactions between different features by approximating the weights for any given co-occurring feature pair, while the main quality of DNN models is their ability to capture higher order feature interactions, which is not feasible for LRs or classic (also known as 2nd order) FM versions. Besides DNN, FMs and LR are among the most popular models in the CTR prediction space. Because of their good performance and relative simplicity, they were selected as our model baselines. Finally, we also selected the more sophisticated DeepFMs, which combine the functionality of FMs and DNN models and are currently considered one of the state-of-the-art approaches for CTR prediction.

While the model is clearly the most important aspect of CTR predictions, some recent studies show that the way we handle features should rightfully get a significant portion of our attention. Since dimensionality of the mentioned categorical features is typically extremely high, a dimensionality reduction approach called the hashing trick was proposed by Weinberger \etal \cite{10.1145/1553374.1553516}. The hashing trick aims to reduce feature dimensionality by using a hashing function to map features from their original space to a smaller feature space. Different from random projections, the hashing trick introduces no additional overhead to store projection matrices. Even more, it actually helps reduce storage needs by reducing feature dimensionality.

Our research builds on the work by He \etal \cite{facebook1} and Zhou \etal \cite{zhou2019resembedding} which suggests that intelligent feature embedding and extraction could increase predictive performance of models and reduce the need for manual feature engineering. The main goal of our research is thus to implement and evaluate whether various feature embedding approaches can indeed improve the efficiency in training and accuracy of predictive models. To achieve this goal, we explore different feature embedding approaches in the context of CTR prediction. While the presented findings are primarily focused around CTR prediction, they could be easily transferred and used in other fields as well, particularly with other types of tabular data classification or regression problems. This research is also a direct continuation of \cite{samothesis}, where we explored performances of similar embedding improvements on a private dataset provided by Zemanta.

\section{methods}

In this section we first present the steps traditionally required for performing model prediction. Next, we present the five embedding modules we developed: the embedding scaling module, the FM embedding module, the embedding encoding module, the NN embedding module and the embedding reweighting module. Finally, we present all of the details of our experimental setup.

\subsection{The general predictive framework}

To adequately investigate different embedding and enrichment approaches, we will first formulate the typical steps that we have to take when using a prediction model. We can break the process into four steps: input, hashing, embedding and prediction. The following subsections briefly describe each step.

\subsubsection{Input}

The first step is serving raw input data to our model. Since raw input data is usually gathered from different sources and tries to capture as much information as possible, it features different data types. In the context of CTR, both numerical as well as categorical features are common. While some predictive models, such as decision trees, are able to learn directly from raw categorical data, others approaches (e.g., FM, DNN, etc.) require numerical inputs. In such cases some kind of feature embedding is usually required to encode categorical data so it can be used for model training and prediction.

\subsubsection{Hashing}

Categorical variables often have very large dimensionality, which can make model training and prediction problematic. A simple and practical solution to this issue is feature hashing, also called the hashing trick~\cite{10.1145/1553374.1553516}. The hashing trick is formally described as follows, for a given feature with value $x_i$, where $i \in N$, implying that feature has dimensionality $\vert N \vert$, define a hashing function $h_{\textrm{hashing}}$:
\begin{align*}
    h_{\textrm{hashing}} \colon N &\to M \\
x_i &\mapsto j,
\end{align*}
where $j \in M$ and $\vert M \vert \ll \vert N \vert$. The hashing trick essentially maps a high dimensional feature into a smaller dimensional space by computing a hash value of the original feature value.

A scenario where two different values get mapped to the same hash value is called a collision. Collisions occur because we are mapping from a larger to a smaller space. However, if we assume a sufficiently large hashing space, the performance loss due to collisions becomes negligible, making this approach very successful in practice. Furthermore, since feature hashing essentially reduces model complexity, it can be considered a form of regularization~\cite{DBLP:journals/corr/abs-1709-03933}. We can perform feature hashing in two ways: either hash each feature separately and treat values from different columns differently, or hash the entire sample. In both cases, we obtain a set of index values, which are forwarded to the embedding layer. The difference is that separate feature hashing avoids cases where a collision occurs between two feature values of different columns that happen to have the same original value. Conversely, hashing the entire sample is faster and easier to implement. Our experiments use the latter example of hashing the entire sample.

\subsubsection{Embedding}

Since predictive models like LR, FM and DeepFM require numerical inputs, categorical data needs to be mapped into a numerical space. We first describe the process of one-hot encoding, which shows how we transform categorical information to numerical vectors and afterwards describe an analogous index based approach that avoids generating large sparse vectors.

The most popular approach for categorical data embedding is called one-hot encoding, which maps categorical values to sparse vectors. We formally describe single-feature one-hot encoding as follows, for a given categorical feature $f$, define set $N$, which contains all possible values of $f$:
\begin{equation*}
    N = \{x_1, x_2, \cdots, x_{\vert N \vert}\}.
\end{equation*}
For example, if our categorical feature is device type, then our possible corresponding set $N$ is $\{\textrm{PC},\ \textrm{laptop},\ \textrm{mobile},\ \textrm{tablet}\}$. Using $N$, we define the following map:
\begin{align*}
    h_{\textrm{one-hot}} \colon N &\to \{0, 1\}^{\vert N \vert} \\
x_i &\mapsto v,
\end{align*}
where $v$ is an $\vert N \vert$-dimensional vector of zeroes, with the $i$-th component equal to $1$. In our example above, this translates to:
\begin{align*}
    h_{\textrm{one-hot}} (\textrm{PC}) &= [1,\ 0,\ 0,\ 0]^T, \\
    h_{\textrm{one-hot}} (\textrm{laptop}) &= [0,\ 1,\ 0,\ 0]^T, \\
    h_{\textrm{one-hot}} (\textrm{mobile}) &= [0,\ 0,\ 1,\ 0]^T, \\
    h_{\textrm{one-hot}} (\textrm{tablet}) &= [0,\ 0,\ 0,\ 1]^T.
\end{align*}
After obtaining such a sparse vector for each categorical feature, we concatenate them. The concatenated vector can then be multiplied with our model weights and used to compute the final prediction. This is however computationally expensive, since we always store all values of each of our one-hot encoded vectors.

To avoid storing large vectors, we can instead simply use the relevant index values to retrieve the relevant model weights directly and avoid multiplication. For example, instead of embedding our feature value $x_i$ into a vector $e_i$ where the $i$-th component equals $1$ and multiplying it with our weight vector w:
\begin{equation*}
    w = [w_0, w_1, \cdots, w_{i-1}, w_i, w_{i+1}, \cdots, w_{\vert N \vert}],
\end{equation*}
which would obtain component $w_i$, we instead use index value $i$ to obtain the component directly. We illustrate the advantage of the the index-based approach in the following example. Let's say we are dealing with a categorical feature device type, with its corresponding value set: 
\begin{equation*}
    N=\{\textrm{PC},\ \textrm{laptop},\ \textrm{mobile},\ \textrm{tablet}\}. 
\end{equation*}
We wish to embed this categorical feature as a 2-dimensional numeric vector. Note that this implies that the weight vector from the above definition actually becomes a weight matrix with dimensions $(2 \times 4)$. An example of such an embedding/weight matrix can be seen below:
\begin{equation*}
    W = \begin{bmatrix}
       0.33 & 0.12 & 2.57 & 3.04  \\
       1.43 & 0.50 & 1.26 & 7.55  \\
     \end{bmatrix}.
\end{equation*}
Obtaining a numerical embedding for the value of ``laptop'' requires the following steps. We first obtain the appropriate one-hot vector:
\begin{equation*}
    h_{\textrm{one-hot}}(\textrm{laptop}) = v_{\textrm{laptop}} = \begin{bmatrix}
    0 \\
    1 \\
    0 \\
    0 \\
    \end{bmatrix},
\end{equation*}
then we proceed to multiply the embedding/weight matrix with the one-hot vector to obtain the final numeric feature embedding:
\begin{equation*}
    W \cdot v_{\textrm{laptop}} = 
    \begin{bmatrix}
       0.33 & 0.12 & 2.57 & 3.04  \\
       1.43 & 0.50 & 1.26 & 7.55  \\
     \end{bmatrix} \cdot \begin{bmatrix}
    0 \\
    1 \\
    0 \\
    0 \\
    \end{bmatrix} = \begin{bmatrix}
       0.12 \\
       0.50 \\
    \end{bmatrix}.
\end{equation*}
The index-based approach simplifies this process by obtaining the index value of the feature value laptop. In this context, the index value indicates the position of value laptop in set $N$:
\begin{equation*}
    h_{\textrm{index}}(\textrm{laptop}) = 1.
\end{equation*}
Finally, the index is used to extract\footnote{Here, the function $\textrm{col}_i(M)$ returns the $i$-th column of matrix $M$.} the embedding directly from the embedding/weight matrix:
\begin{equation*}
    \textrm{col}_1(W) = 
    \textrm{col}_1 \begin{pmatrix}\begin{bmatrix}
       0.33 & 0.12 & 2.57 & 3.04  \\
       1.43 & 0.50 & 1.26 & 7.55  \\
     \end{bmatrix}\end{pmatrix} = \begin{bmatrix}
       0.12 \\
       0.50 \\
    \end{bmatrix}.
\end{equation*}
Not generating the one-hot vectors is an important optimization, since these vectors are typically very large, due to the high dimensionality of categorical features.

\subsubsection{Prediction}
\label{section:actfun}

%We selected LR and FM as our baseline models. These baselines were selected because they are relatively simple and historically achieve good performance in CTR prediction. The following subsections provide a basic theoretical description of both models.

We selected logistic regression (LR), factorization machines (FM) and deep factorization machines (DeepFM) as our baseline models. The following subsections provide a basic theoretical description of the mentioned models.

\begin{figure}[ht]
    \begin{center}
        \includegraphics[width=12cm]{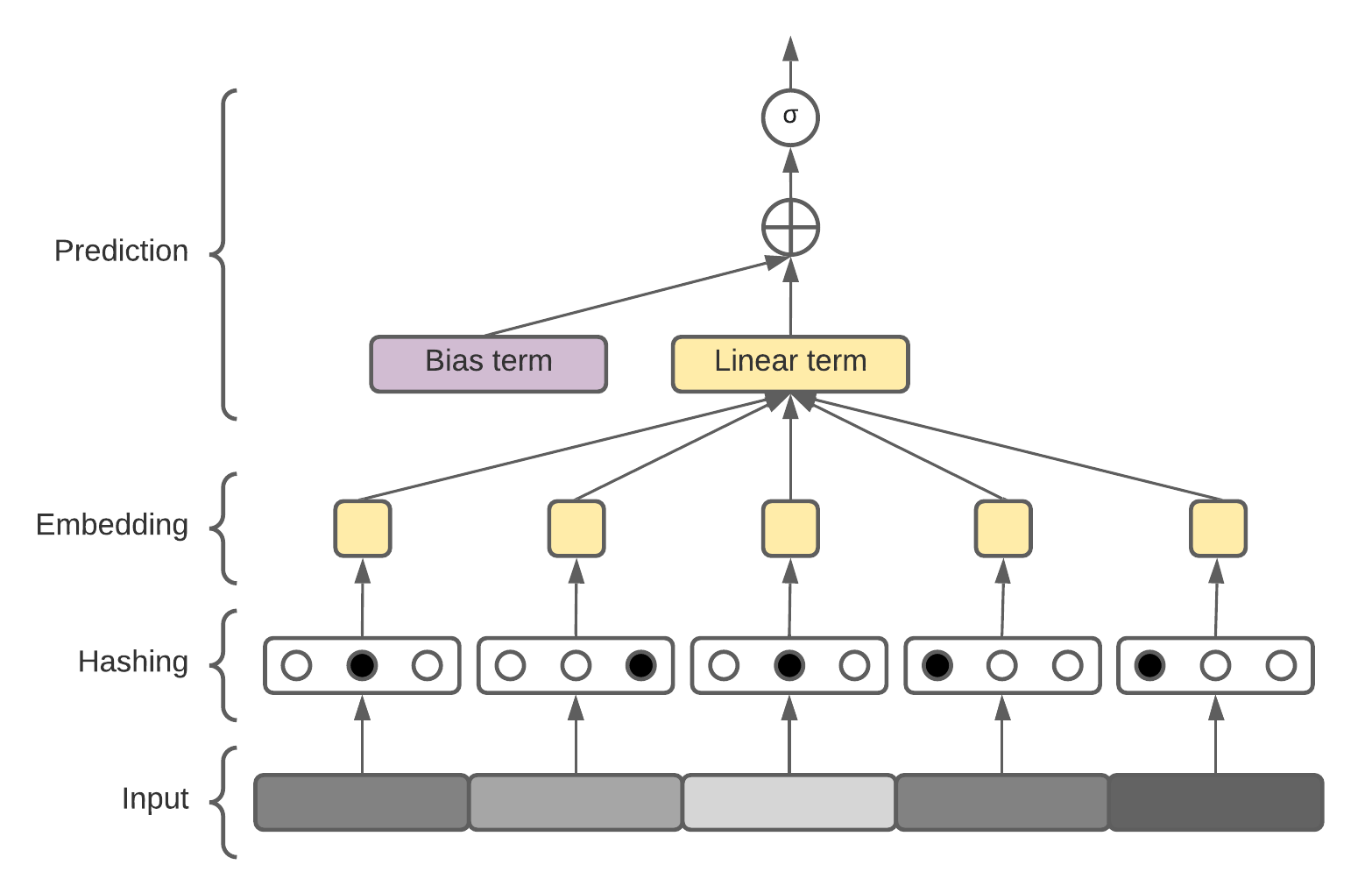}
    \end{center}
\caption{\textbf{A visual representation of the LR model.} LR model performs prediction by computing the sigmoid activation of the summation of its bias and linear terms. Linear embeddings and subsequent linear term are colored yellow, while the bias term is colored purple.}
\label{lr_model}
\end{figure}

LR is a type of predictive analysis that attempts to explain the relationship between one dependent binary variable and one or more independent variables. When performing LR on $m$-dimensional real vectors, the model consists of two components:
\begin{itemize}
    \item an $m$-dimensional weight vector $\theta$,
    \item a bias $\theta_0$.
\end{itemize}
For a given sample $x \in \mathbb{R}^m$, our model prediction $f(x)$ equals:
\begin{equation*}
 f(x) = \sigma(\theta_0 + \theta_1x_1 + \theta_2x_2 + \dots \theta_mx_m) = \sigma(\theta_0 + \sum^m_{i=1} \theta_ix_i),  
\end{equation*}
where $\sigma$ is the \textit{logistic function}. The logistic function is a member of the sigmoid function family and is defined by the formula:
\begin{equation*}
\sigma(x) = \frac{1}{1 + e^{-x}}.
\end{equation*}
LR is essentially a linear model that outputs the probability of a positive outcome for a binary event given a set of dependent real variables. It is attractive in the CTR prediction context due to its simplicity, training and prediction speeds and decent performance~\cite{41159}. The LR model is visualized in Figure~\ref{lr_model}.

\begin{figure}[ht]
    \begin{center}
        \includegraphics[width=12cm]{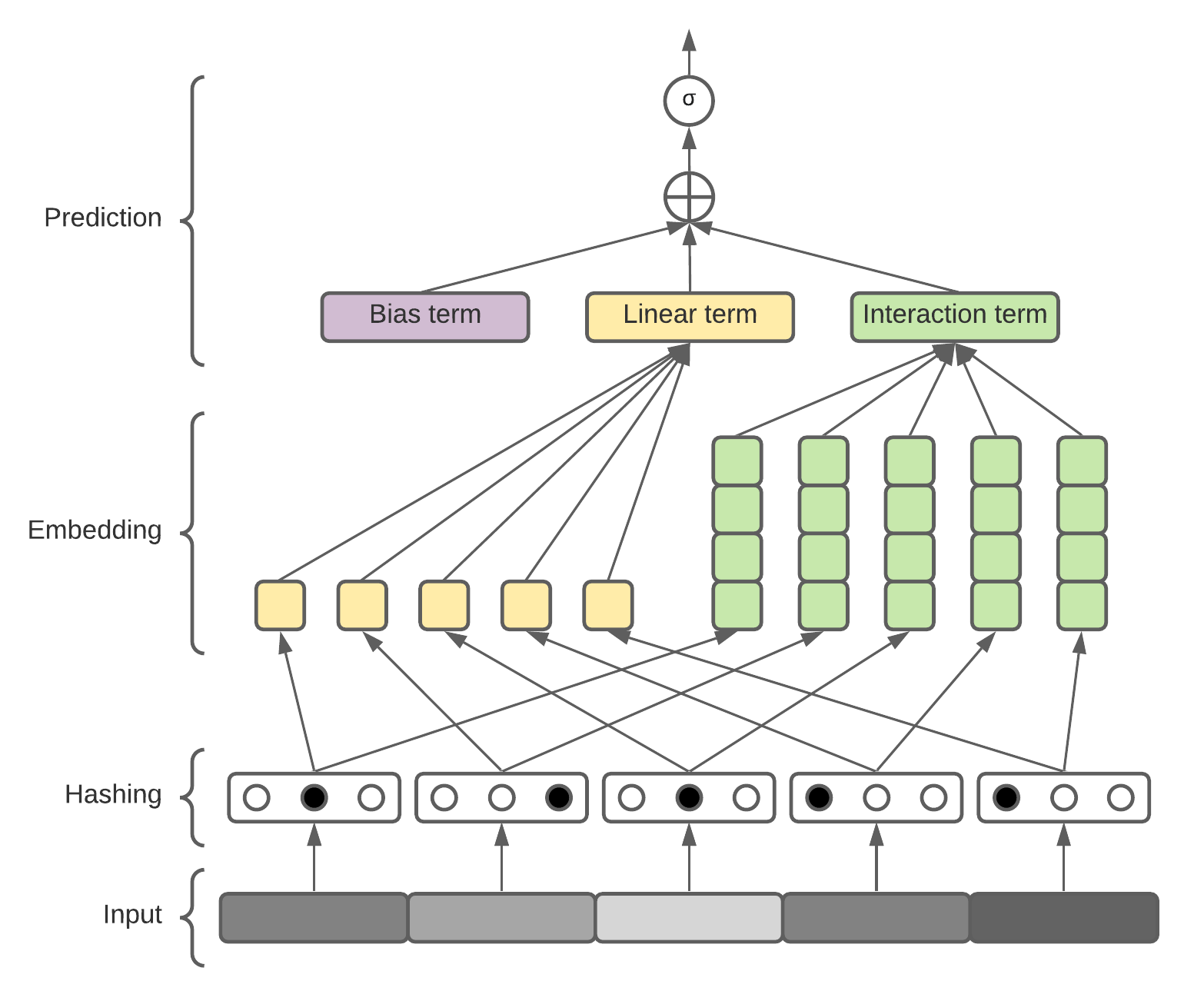}
    \end{center}
\caption{\textbf{A visual representation of the FM model.} FM model performs prediction by computing the sigmoid activation of the summation of its bias, linear and interactions terms. Ignoring the interaction embeddings, the model downgrades to classic LR. Interaction embedding and subsequent interaction term are colored green.}
\label{fm_model}
\end{figure}

FMs attempt to capture interactions between features by using factorized parameters. They can be utilized to model any order of feature interactions, although second order interactions are the most common. When performing prediction on $m$-dimensional real vectors via a second order FM, we require three components: 
\begin{itemize}
    \item an $(m \times k)$-dimensional factorized interaction matrix $V$; here $k$ denotes the size of the interaction vectors,
    \item an $m$-dimensional weight vector $\theta$,
    \item a bias $\theta_0$.
\end{itemize}
For a given sample $x \in \mathbb{R}^m$, our model prediction $f(x)$ equals:

\begin{equation*}
f(x) = \sigma(\theta_0 + \sum^m_{i=1} \theta_ix_i + \sum^m_{i=1}\sum^m_{i<j}\langle v_i, v_j \rangle x_i x_j).
\end{equation*}

Notably, the FM prediction on an $m$-dimensional vector is equal to LR, with the addition of the interaction term. The interaction term is used to approximate all second order feature interactions by computing scalar products between their respective latent vectors. The latent vectors are rows in matrix $V$, so the value of feature $x_i$ corresponds to the $i$-th row in V, denoted as vector $v_i$. The above formalization also illustrates why FM are well suited for prediction problems where data is high-dimensional and sparse. If our $m$-dimensional vector $x$ is sparse, only a small number of non-zero feature combinations need to be computed. The FM model is visualized in Figure~\ref{fm_model}.

\begin{figure}[ht]
    \begin{center}
        \includegraphics[width=12cm]{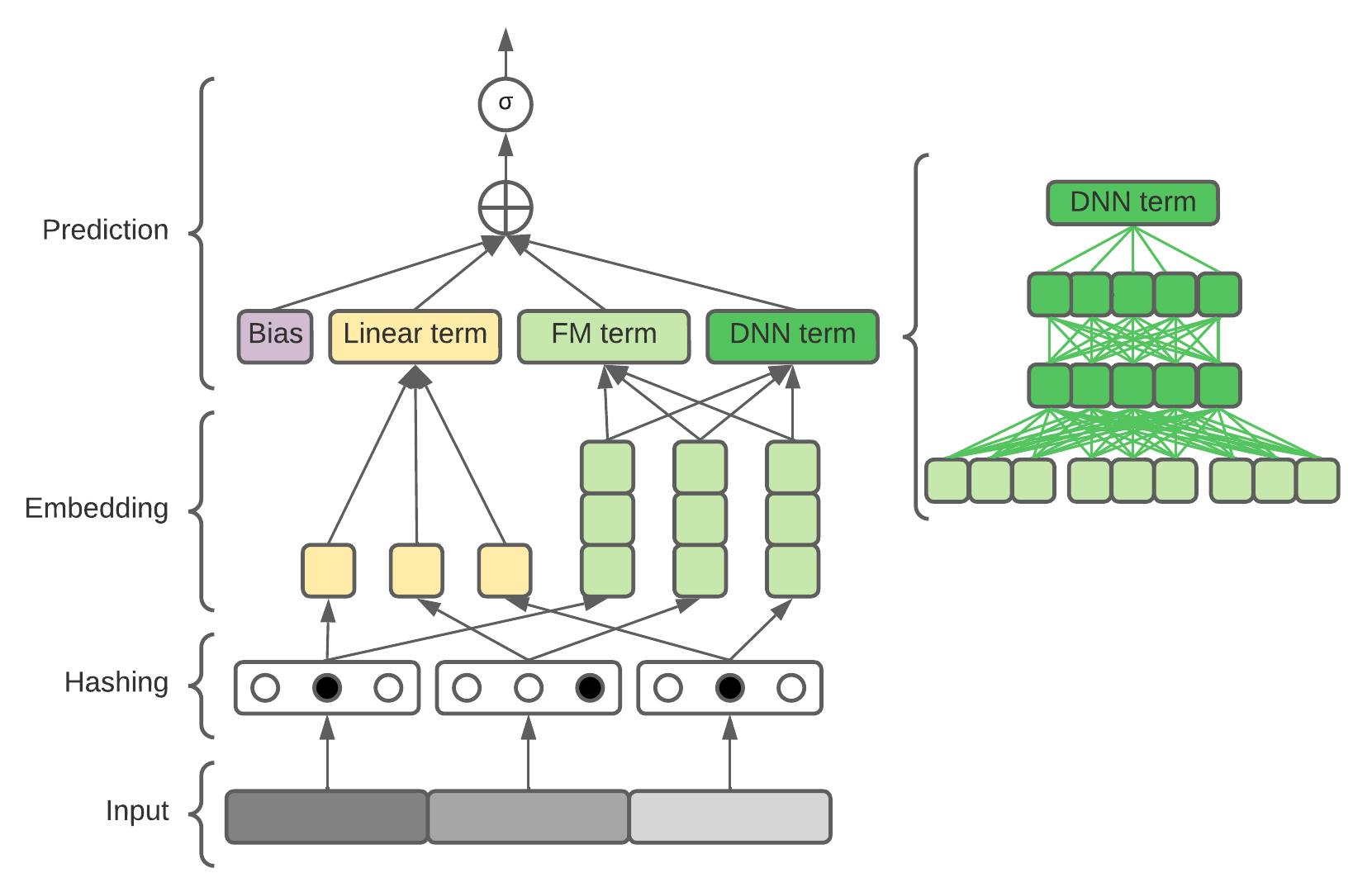}
    \end{center}
\caption{\textbf{A visual representation of the DeepFM model.} DeepFM model performs prediction by computing the sigmoid activation of the summation of its bias, linear, interaction and DNN terms. The DNN term is computed by using the same set of embeddings used to compute the FM term.}
\label{deepfm_model2}
\end{figure}

Our final model are deep factorization machines or DeepFMs. Proposed by Guo \etal in \cite{guo2017deepfm}, they improve on the performance of FMs by adding an additional DNN term to the final prediction. When performing prediction on $m$-dimensional real vectors via a DeepFM, we require:
\begin{itemize}
    \item an $(m \times k)$-dimensional factorized interaction matrix $V$; here $k$ is a hyperparameter denoting the size of the interaction vectors,
    \item an $m$-dimensional weight vector $\theta$,
    \item a bias $\theta_0$,
    \item a neural network used to compute the DNN term; neural network size is chosen as a hyperparameter.
\end{itemize}

For a given sample $x \in \mathbb{R}^m$, our model prediction $f(x)$ equals:

\begin{equation*}
f(x) = \sigma(\theta_0 + \sum^m_{i=1} \theta_ix_i + \sum^m_{i=1}\sum^m_{i<j}\langle v_i, v_j \rangle x_i x_j + y_{DNN}).
\end{equation*}

Notably, the DeepFM prediction on an $m$-dimensional vector is equal to FM, with the addition of the DNN term $y_{DNN}$. The DNN term is used to capture high-order interactions of the input embeddings by feeding them through a series of fully connected neural network layers. 

\subsubsection{Performing feature embedding}

Considering the described prediction process, our objective is to devise an additional step, embedding+. The embedding+ step is included among the prediction steps and aims to produce an improved embedding which in turn results in improved predictions. The location of the embedding+ step is technically arbitrary and based on the embedding approach in question, but we primarily focus on approaches that perform the step between embedding and prediction steps. Notably, the embedding+ step differs from other established embedding techniques, like word2vec~\cite{https://doi.org/10.48550/arxiv.1301.3781}, because it is a direct part of the end-to-end training process. Furthermore, since it is trained alongside the rest of the model, its performance is dynamic. 

\subsection{Embedding modules}

This section describes the feature embedding modules we implemented in order to try improving the predictive performance of the baseline models. Visualizations maintain the previously introduced color coding, linear embeddings and related terms are visualized with a yellow color, while the interaction embeddings are visualized with a green color. Module architecture and resulting embedding vectors are presented in blue.

\subsubsection{The embedding scaling module}
\label{section:scaling}

\begin{figure}
    \centering
    \subfloat{{\includegraphics[width=5cm]{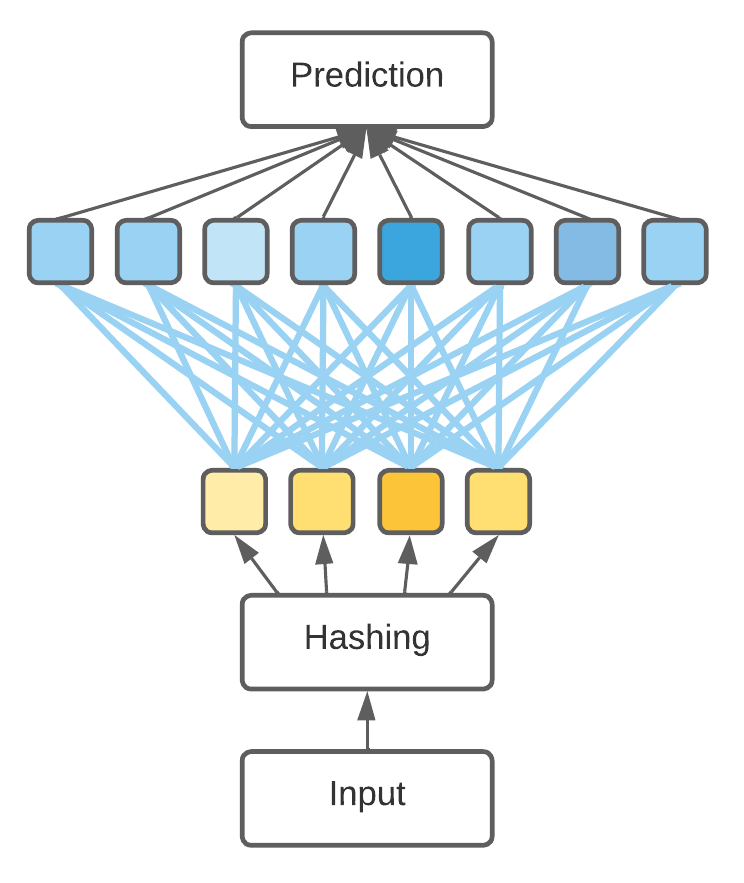} }}%
    \qquad
    \subfloat{{\includegraphics[width=5cm]{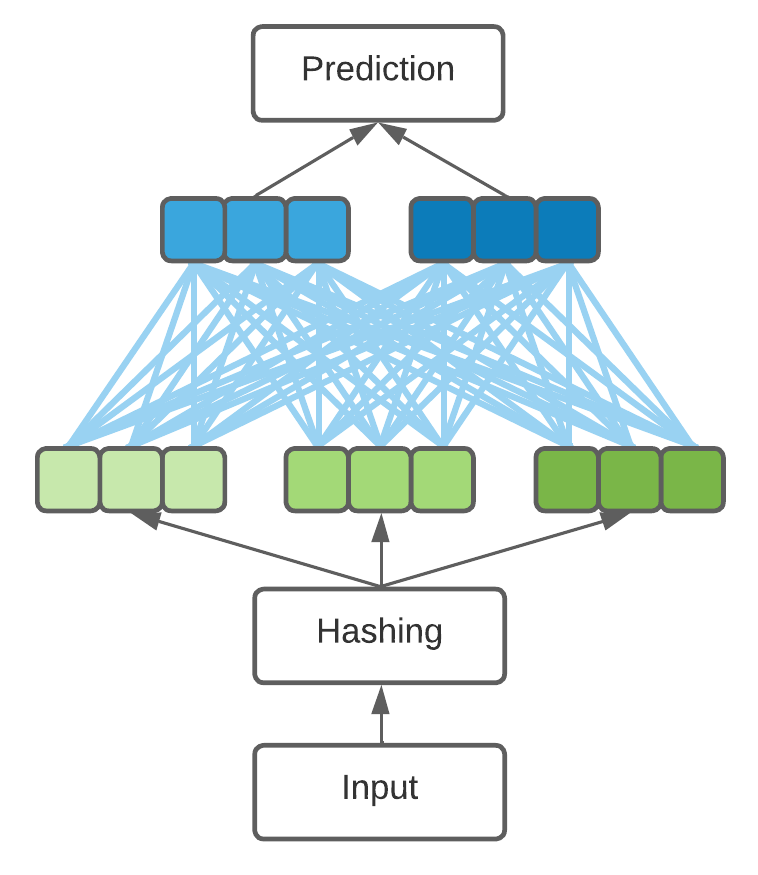} }}%
    \caption{\textbf{A visual representation of the embedding scaling module.} Embedding dimension can be either upscaled (left) or downscaled (right).}%
    \label{scaling}
\end{figure}

The embedding scaling module aims to improve final model's predictive performance by feeding the dense embeddings into a fully connected neural network. The neural network has $H$ hidden layers, where $H \in \mathbb{N}_0$ is a tunable hyperparameter. Each hidden layer as well as the output layer have $\round{F \cdot S}$ neurons, where $F$ is the number of model features and thus the size of the original embedding vector and $S \in \mathbb{R}$ is the scaling hyperparameter. Notably, we use the parameter $S$ to either upscale or downscale the dimension of the original embedding vector. Each approach has its own motivation: upscaling seeks to increase the model's expressiveness, while downscaling seeks to reduce the model's overfitting. The (linear) embedding scaling process with $H$ hidden layers and scaling factor $S$ is performed on a dataset with $F$ categorical features. For a given data sample with $F$ categorical features, our embedding layer returns the following set of dense embeddings. Each set element is a real number corresponding to the value of the respective categorical feature:
\begin{equation*}
    e = \{e_1,\ e_2,\ \dotsc, e_F\}.
\end{equation*}
We define $H+1$ matrices: $W_0$, $W_1, \dotsc, W_H$. Matrix $W_0$ has dimensions $(F \times R)$, where $R = \round{F \cdot S}$ represents the size of the final rescaled embedding vector. Matrices $W_1, W_2, \cdots, W_H$ have dimensions $(R \times R)$. For each matrix, we also define corresponding bias vectors $b_0, b_1, \cdots, b_H$, which get added to the result of each performed matrix multiplication. All bias vectors are of size $R$. Our set of dense embeddings is transformed into a single vector:
\begin{equation*}
    e' = [e_1,\ e_2,\ \dotsc, e_F],
\end{equation*}
which is first multiplied with matrix $W_0$ and added to bias vector $b_0$. We also apply an activation function, denoted below as $\sigma(.)$, over each element of the result, to break linearity. This produces the first scaled embedding vector:
\begin{equation*}
    w_0 = \sigma(W_0 \cdot e' + b_0).
\end{equation*}
The scaled embedding vector $w_0$ is afterwards multiplied with the remaining matrices in the following manner:
\begin{align*}
    w_1 &= \sigma(W_1 \cdot w_0 + b_1), \\
    w_2 &= \sigma(W_2 \cdot w_1 + b_2), \\
    &\cdots, \\
    w_H &= \sigma(W_H \cdot w_{H-1} + b_H).
\end{align*}
We consider the final output vector $w_H$ as our rescaled embedding and forward it to the prediction layer. Figure~\ref{scaling} shows two examples of embedding scaling. 

The embedding scaling module contains $R \cdot K((F \cdot K+1) + (R \cdot K+1)(H-1))$ parameters, where $F$ equals the number of features, $K$ equals the size of the embedding vectors, $H$ equals the number of hidden layers and $R = \round{F \cdot S}$ is the new rescaled number of features. Notably, the number of parameters is considerably lower than the total dimensionality of the categorical feature space, which implies that the size of the embedding scaling module is practically negligible compared to the baseline model embedding layer size. The embedding scaling module also has a special interaction with the DeepFM prediction model. Since the size of the DeepFM's neural network component is related to the number of model features, it is influenced by the rescaling operation of the module. This means that the upscaling operation of the module results in a larger DeepFM neural network and vice versa for downscaling.

\subsubsection{The FM embedding module}
The FM embedding module aims to transform the set of original interaction embedding vectors into a different set and use it to make the final prediction by computing all possible combinations of scalar products between vectors. While the classic FM model trains the set of interaction vectors that correspond to the samples' categorical features, the FM embedding module instead directly focuses on the interaction vector components. 

The FM embedding process with original interaction vector of size $K$ and new interaction vector of size $C$ is performed on a dataset with $F$ categorical features. For a given data sample with $F$ categorical features, our embedding layer returns the following set of interaction vectors:
\begin{equation*}
    e = \{ \vec{e}_1, \vec{e}_2, \dotsc, \vec{e}_F \} = \begin{Bmatrix}\begin{bmatrix}e_{11} \\ e_{21}\\ \vdots \\ e_{K1}\end{bmatrix}, \begin{bmatrix}e_{12} \\ e_{22}\\ \vdots \\ e_{K2}\end{bmatrix}, \dotsc, \begin{bmatrix}e_{1F} \\ e_{2F}\\ \vdots \\ e_{KF}\end{bmatrix}\end{Bmatrix}.
\end{equation*}
Instead of computing the interaction term as the sum of scalar products between all possible pairs of vectors from set $e$, we transform $e$ into a single dense vector $e'$ and treat it as a data sample:
\begin{equation*}
    e' = [e_{11}, e_{21}, \dotsc, e_{K1}, \dotsc, e_{KF}].
\end{equation*}
We now aim to model interactions between features of sample $e'$. To achieve this, we define a trainable vector set $v$, which contains $N = KF$ vectors of size $C$: 
\begin{equation*}
    v = \{ \vec{v}_1, \vec{v}_2, \dotsc, \vec{v}_N \} = \begin{Bmatrix}\begin{bmatrix}v_{11} \\ v_{21}\\ \vdots \\ v_{C1}\end{bmatrix}, \begin{bmatrix}v_{12} \\ v_{22}\\ \vdots \\ v_{C2}\end{bmatrix}, \dotsc, \begin{bmatrix}v_{1N} \\ v_{2N}\\ \vdots \\ v_{CN}\end{bmatrix}\end{Bmatrix}.
\end{equation*}
We treat these vectors as new interaction vectors used to approximate interactions between features of our data sample $e'$. We assign each component of our data sample a separate trainable interaction vector. Each component from $e'$ gets multiplied element-wise to its corresponding vector from set $v$. This results in a new set of interaction embedding vectors:
\begin{equation*}
    v' = \{ e_1 \odot \vec{v}_1,\ e_2 \odot \vec{v}_2,\ \dotsc,\ e_{KF} \odot \vec{v}_N \}.
\end{equation*}
The new embedding vector set gets forwarded to the prediction layer, where a prediction is performed in the same way as with classic FMs. Recalling the original FM equation, the new interaction term $y_{\mathrm{interaction}}$ calculation can be reformulated as follows:
\begin{equation*}
    y_{\mathrm{interaction}} = \sum^{N}_{i=1}\sum^N_{i<j} \langle v_i, v_j \rangle e'_i e'_j.
\end{equation*}

\begin{figure}
    \begin{center}
        \includegraphics[width=4.5cm]{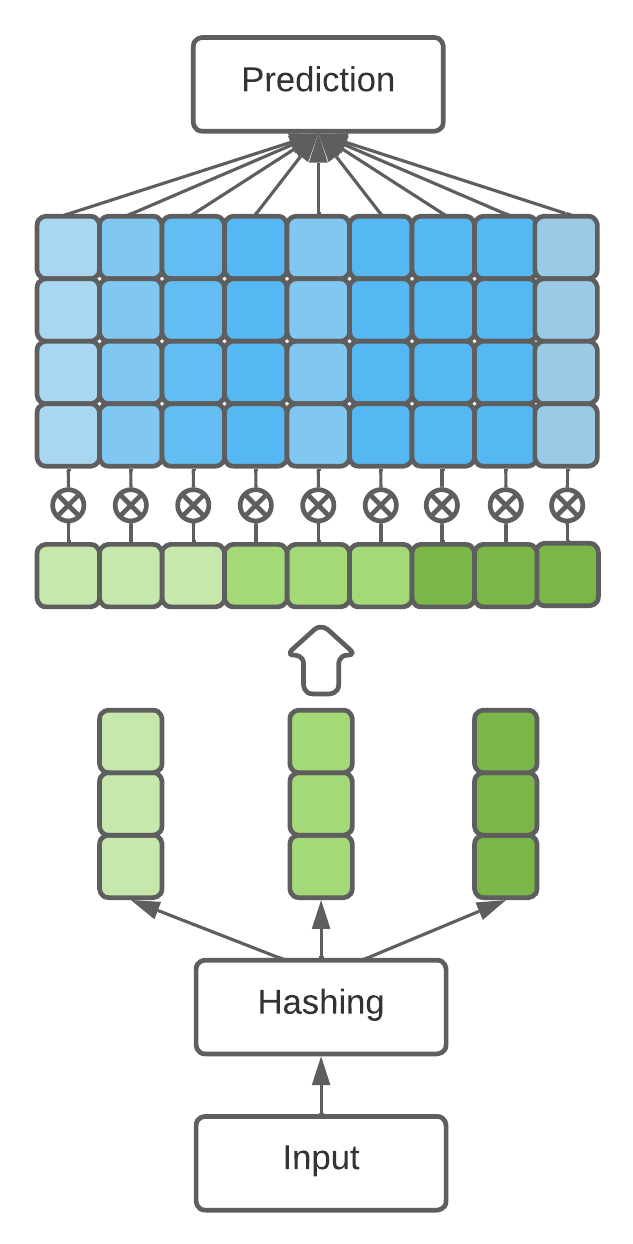}
    \end{center}
\caption{\textbf{A visual representation of the FM embedding module.} Embedding via FM is performed in two steps: first, extract the basic embedding, and second, multiply each original embedding weight with a separate embedding vector in an element-wise fashion. The resulting set of embedding vectors is then used for the final prediction.}
\label{metafm}
\end{figure}

The FM embedding module contains $F \cdot K \cdot C$ parameters, where $F$ equals the number of features, $K$ equals the size of the original embedding vectors and $C$ equals the size of the new embedding vectors. A visualization of this embedding module can be seen in Figure~\ref{metafm}. Similarly to the embedding scaling module, the FM embedding module also influences the size of the DeepFM's neural network component. In this case, the result is always an increase in component parameters, since the effective number of features is increased from $F$ to $F \cdot K$.

\subsubsection{The embedding encoding module}
\begin{figure}[ht]
    \centering
    \subfloat{{\includegraphics[width=2.4cm]{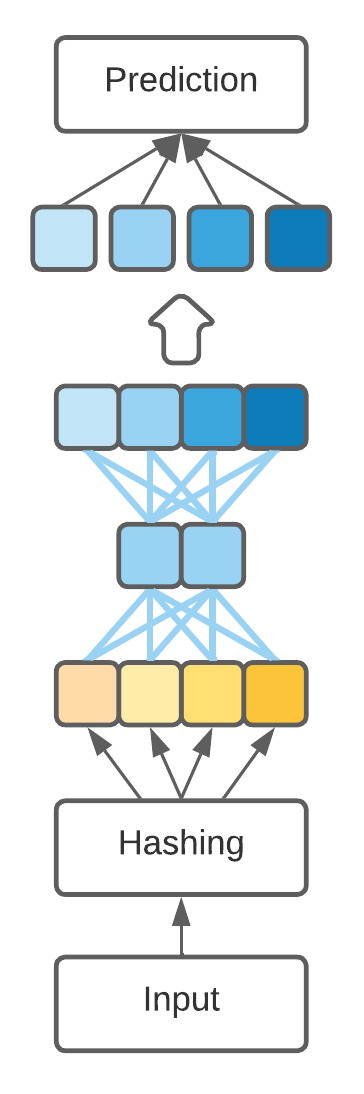} }}%
    \qquad
    \qquad
    \subfloat{{\includegraphics[width=4cm]{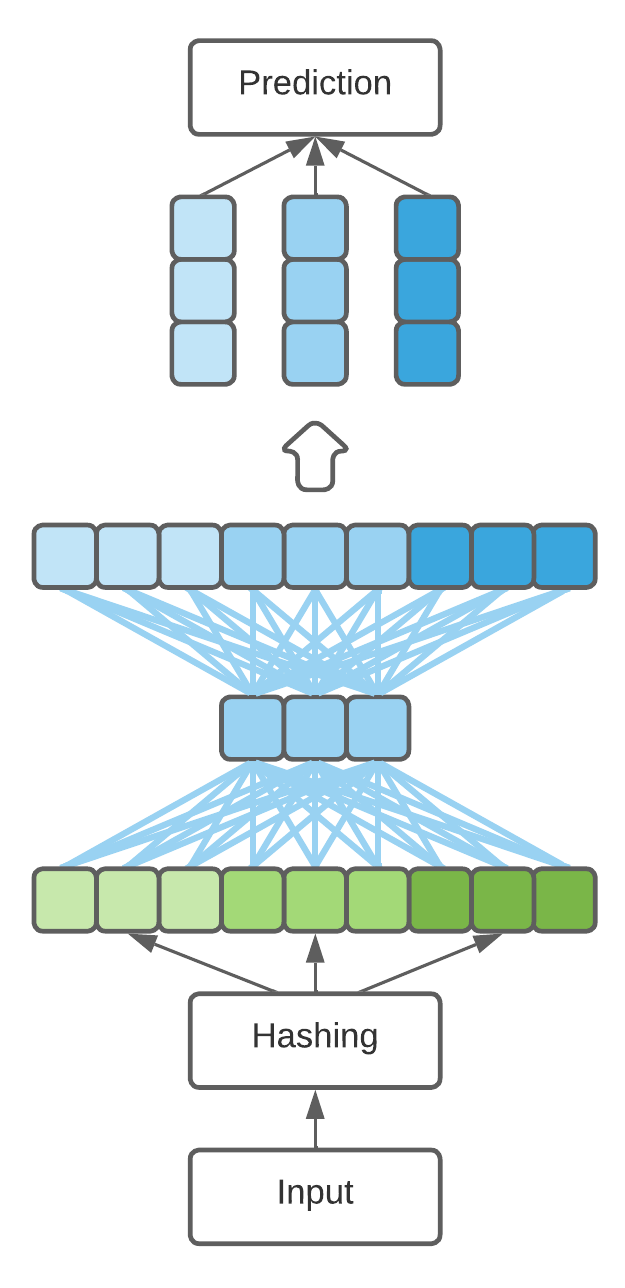} }}%
    \caption{\textbf{A visual representation of the embedding encoding module.} The embedding is concatenated into a single vector, which is fed into the neural network with a narrow hidden layer. Afterwards, it is reconstructed into a vector of the same size as the input and reshaped into separate feature embeddings.}%
    \label{squeeze}
\end{figure}

The embedding encoding module aims to improve embedding quality by feeding the existing embedding vectors through a neural network with a narrow hidden layer and afterwards reconstructing the hidden layer output to the original input size. We adopt this dimensionality reduction approach from classic autoencoders~\cite{bank2021autoencoders}. The embedding encoding procedure for linear as well as interaction embeddings can be seen in Figure~\ref{squeeze}. 

The embedding encoding process with shrinking factor $S \in [1, \infty)$ is performed on a dataset with $F$ categorical features. For a given data sample with $F$ categorical features, our embedding layer returns a set of dense embeddings. Each set element is a real vector of size $K$, corresponding to the value of the respective categorical feature:
\begin{equation*}
    e = \{e_1,\ e_2,\ \dotsc, e_F\}.
\end{equation*}
To perform our embedding encoding and decoding steps, we define two matrices, $W_{\textrm{contract}}, W_{\textrm{expand}}$, with dimensions $(F \times E)$ and $(E \times F)$, where $E = \round{F / S}$ equals the size of the narrow representation vector. Identically to section \ref{section:scaling}, corresponding bias vectors $b_{\textrm{contract}}, b_{\textrm{expand}}$ are defined for each matrix. Our set of dense embeddings is concatenated into a single vector:
\begin{equation*}
    e' = [e_1,\ e_2,\ \dotsc, e_F],
\end{equation*}
and afterwards fed into the network. The process is again similar to \ref{section:scaling} and likewise utilizes activation functions to break linearity:
\begin{align*}
    w_0 &= \sigma(W_{\textrm{contract}} \cdot e' + b_{\textrm{contract}}), \\
    w_1 &= \sigma(W_{\textrm{expand}} \cdot w_0 + b_{\textrm{expand}}).
\end{align*}
We consider the final output vector $w_1$ as our improved embedding and forward it to the prediction layer. Description of hidden layers is omitted for brevity. An arbitrary number of hidden layers can be included both before and after the narrow hidden layer. Description of such hidden layers is found in section~\ref{section:scaling}. 

%The embedding encoding module contains $F(H+1) + H(F+1) + X$ parameters, where $F$ equals the size of the original embedding vector, $H$ is the size of the narrow hidden layer vector and $X$ is the number of parameters in the accompanying hidden layer topology (model description above assumes $X=0$).

The embedding encoding module contains $EK (FK (EK + FK) + (H-1)(EK + 1) ))$, where $F$ equals the number of features, $K$ equals the embedding vector size, $H$ equals the number of hidden layers and $E = \round{F / S}$ is the size of the of the narrow representation vector. A visualization of this embedding module can be seen in Figure~\ref{squeeze}. Notably, the embedding encoding module differs from embedding scaling and FM embedding because it preserves the effective number of features from the original model. 

\subsubsection{The NN embedding module}
The NN embedding module works by accepting both linear feature embeddings as well as factorized interaction embeddings as its input. After obtaining both types of embedding vectors, we concatenate them and feed the concatenated vector into a fully connected neural network. The motivation behind the NN embedding module is sharing the information between linear and interaction embeddings. Each layer of the module's neural network has $F \cdot (K+1)$ neurons, which corresponds to the size of the concatenated vector. 

\begin{figure}[ht]
    \begin{center}
        \includegraphics[width=3cm]{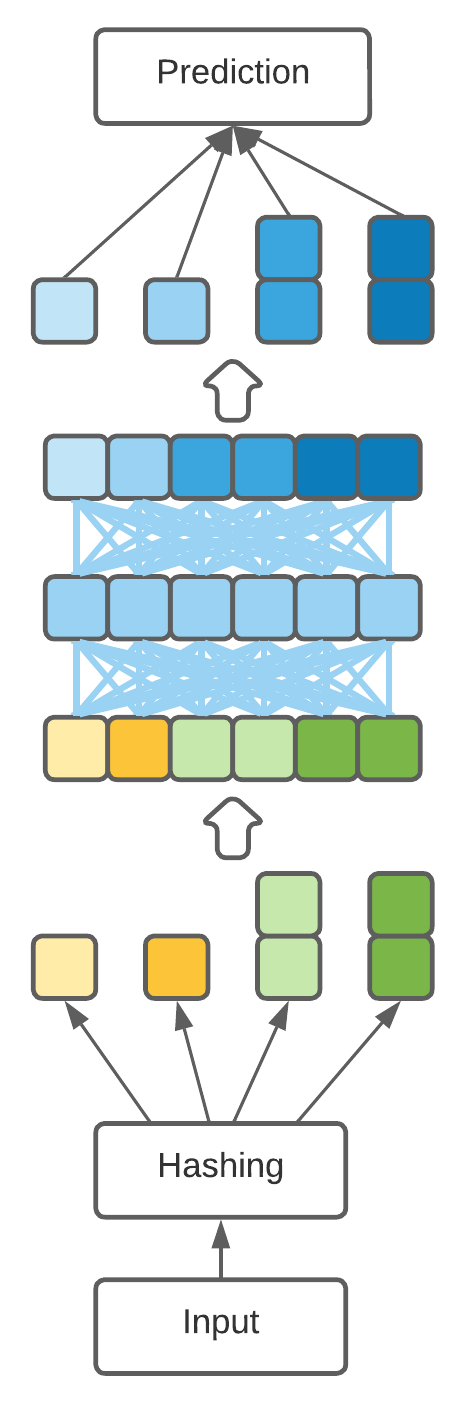}
    \end{center}
\caption{\textbf{A visual representation of the NN embedding module.} NN embedding is performed in three steps: first, extract basic linear and interaction embeddings, second, construct a single vector and feed it into the network, and third, reconstruct the network output into linear and interaction embeddings and use them to perform the final prediction.}
\label{netfm}
\end{figure}

The neural network embedding process with $H \in \mathbb{N}_0$ hidden layers and interaction vector size $K$ is performed on a dataset with $F$ categorical features. For a given data sample with $F$ categorical features, our embedding layer returns the following sets of dense embeddings. Elements from $e_{\textrm{linear}}$ are real numbers, while elements from $e_{\textrm{interaction}}$ are real vectors of size $K$:
\begin{align*}
    e_{\textrm{linear}} &= \{ e_1, e_2, \dotsc, e_F \}, \\
    e_{\textrm{interaction}} &= \{ \vec{e}_1, \vec{e}_2, \dotsc, \vec{e}_F \} = \begin{Bmatrix}\begin{bmatrix}e_{11} \\ e_{21}\\ \vdots \\ e_{K1}\end{bmatrix}, \begin{bmatrix}e_{12} \\ e_{22}\\ \vdots \\ e_{K2}\end{bmatrix}, \dotsc, \begin{bmatrix}e_{1F} \\ e_{2F}\\ \vdots \\ e_{KF}\end{bmatrix}\end{Bmatrix}.
\end{align*}
We define $H+1$ matrices, $W_0, W_1, \dotsc, W_H$, with dimensions $(N \times N)$, where $N = F (K+1)$. We also define bias vectors $b_0, b_1, \dotsc, b_H$ corresponding to each matrix. All elements from our embedding sets are concatenated into a single vector:
\begin{equation*}
    e' = [e_1, e_2, \dotsc, e_F, e_{11}, e_{21}, \dotsc, e_{KF}],
\end{equation*}
and afterwards fed into the network as follows:
\begin{align*}
    w_0 &= \sigma(W_0 \cdot e' + b_0), \\
    w_1 &= \sigma(W_1 \cdot w_0 + b_1), \\
    w_2 &= \sigma(W_2 \cdot w_1 + b_2), \\
    &\cdots, \\
    w_H &= \sigma(W_H \cdot w_{H-1} + b_H).
\end{align*}
We consider the final output vector $w_H$ as our improved embedding. Next, we perform vector reshaping to obtain appropriate embedding sets and forward them to the prediction layer. 

The NN embedding module contains $H(F^2(K+1)^2 + F(K+1))$ parameters, where $F$ equals the number of model features, $K$ is the size of the interaction vectors and $H$ is the number of hidden layers in the module network. Figure \ref{netfm} shows an example module with $F=2$, $K=2$ and $H=1$.

\subsubsection{The embedding reweighting module}
Embedding reweighting aims to improve the model's performance by assigning each feature a weight based on the entire dense embedding vector. These weights range between $0$ and $1$ and predict the degree of relevance each feature will have when making the final prediction. Afterwards, each embedding is scaled by its respective weight and served to the final model. To practically obtain the weight vector for a specific sample, we feed it to a fully connected neural network with no hidden layers and $F$ output neurons, where $F$ is the number of features and a sigmoid activation function. This approach seeks to minimize the influence of noisy features.

\begin{figure}[ht]
    \centering
    \subfloat{{\includegraphics[width=2cm]{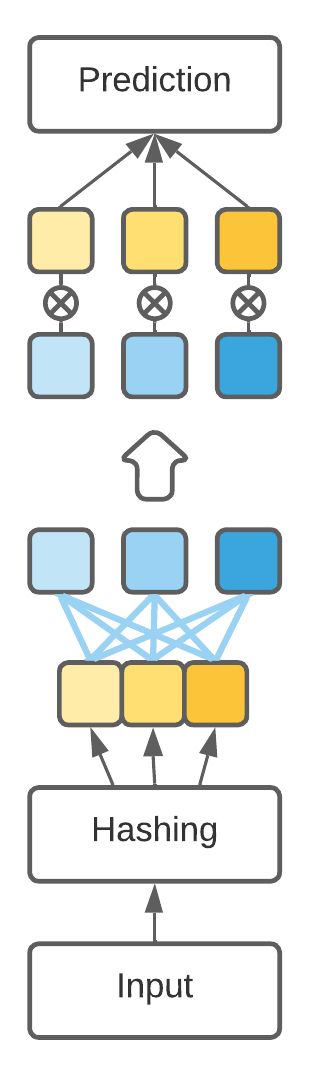} }}%
    \qquad
    \qquad
    \subfloat{{\includegraphics[width=4cm]{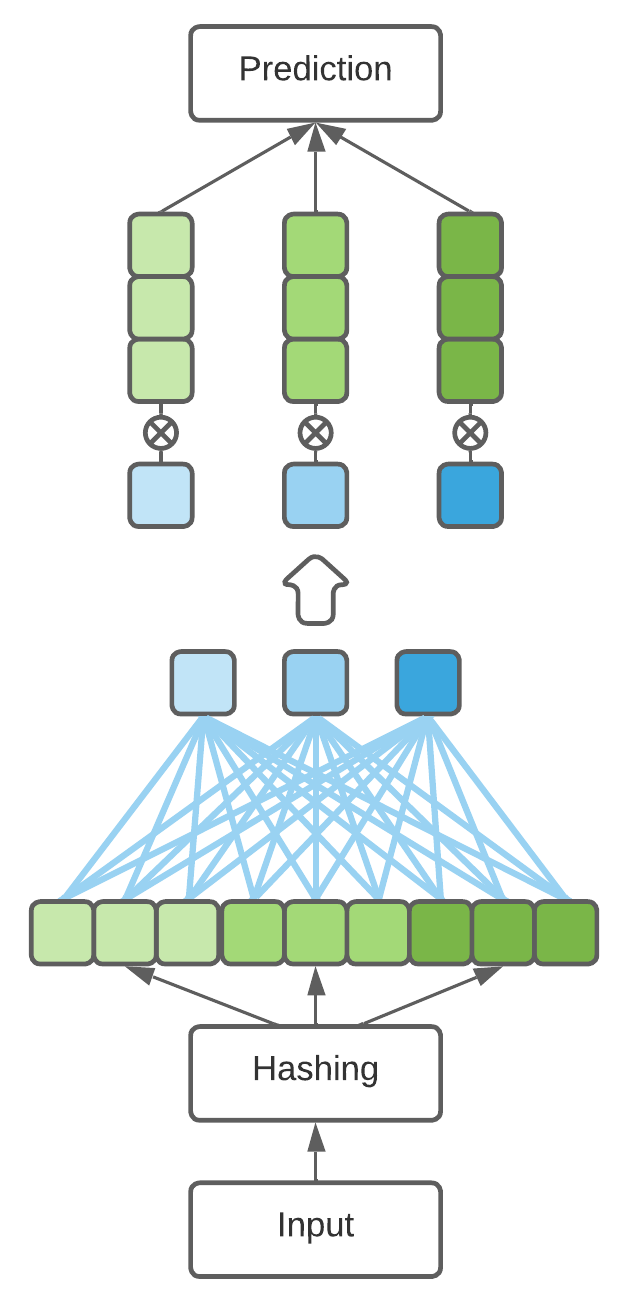} }}%
    \caption{\textbf{A visual representation of the interaction embedding reweighting module.} Feature reweighting is performed in three steps: first, extract the original embedding, second, compute the weight vector, and third, perform element-wise multiplication of the original embedding and the weight vector and use the obtained vector to perform the final prediction.}%
    \label{weight}
\end{figure}

The (interaction) embedding reweighting process with interaction vector size $K$ is performed on a dataset with $F$ categorical features. For a given data sample with $F$ categorical features, our embedding layer returns the following set of interaction vectors:
\begin{equation*}
    e = \{ \vec{e}_1, \vec{e}_2, \dotsc, \vec{e}_F \} = \begin{Bmatrix}\begin{bmatrix}e_{11} \\ e_{21}\\ \vdots \\ e_{K1}\end{bmatrix}, \begin{bmatrix}e_{12} \\ e_{22}\\ \vdots \\ e_{K2}\end{bmatrix}, \dotsc, \begin{bmatrix}e_{1F} \\ e_{2F}\\ \vdots \\ e_{KF}\end{bmatrix}\end{Bmatrix}.
\end{equation*}
To compute our weight vector, we define matrix $W$ with dimensions $(KF \times F)$ and a corresponding bias vector $b$ of size $F$. We concatenate all vectors from our embedding set into a single vector:
\begin{equation*}
    e' = [e_1, e_2, \dotsc, e_F, e_{11}, e_{21}, \dotsc, e_{KF}],
\end{equation*}
and use it to compute a weight vector as follows:
\begin{equation*}
    w = \sigma(W \cdot e' + b).
\end{equation*}

Here, $\sigma(.)$ denotes the logistic activation function (description found in section \ref{section:actfun}). We associate each component of the weight vector $w = [w_1, w_2, \dotsc, w_F]$ with the respective interaction vector from the original embedding set $e$. Associated weights and embedding vectors are multiplied together, resulting in an improved embedding set:

\begin{equation*}
    e_{\textrm{reweighted}} = \{ w_1 \odot \vec{e}_1,\ w_2 \odot \vec{e}_2,\ \dotsc,\ w_F \odot \vec{e}_F \},
\end{equation*}
which is forwarded to the prediction layer. The process above describes reweighting of interaction embeddings. Reweighting of linear embeddings is performed in a similar manner. Both approaches are visualized in Figure~\ref{weight}. 

The embedding reweighting module contains $F(KF + 1)$ parameters, where $F$ equals the number of model features and $K$ is the size of the interaction vectors.

\subsubsection{A summary of embedding modules}

We developed 5 embedding modules, each with its own characteristics. Table~\ref{tab:params} contains an overview of all implemented modules. 

\begin{table}[ht]
\centering
\scalebox{1}{  % previously 0.82
\begin{tabular}{l c c}
 \toprule
 Module name & Target embedding & Module size \\
 \midrule
 Embedding scaling module & Either/Both &  $R \cdot K((F \cdot K+1) + (R \cdot K+1)(H-1))$ \\
 FM embedding module & Interaction &  $FKC$ \\
 Embedding encoding module & Either/Both &  $EK (FK (EK + FK) + (H-1)(EK + 1) ))$ \\
 NN embedding module & Both &  $H(F^2(K+1)^2 + F(K+1))$ \\
 Embedding reweighting module & Either/Both &  $F(KF+1)$ \\
 \bottomrule
\end{tabular}
}
\caption{\textbf{A summary of embedding modules}. Each parameter is described in the respective module section. The FM embedding module can only be applied to interaction embeddings. The NN embedding module requires both linear and interaction embeddings. Others can be applied to linear or interaction embedding or to both of them. The equations that denote the module's size are explained in detail in sections describing the modules.}
\label{tab:params}
\end{table}

\subsection{Experimental setup}
As mentioned in the prior sections, our main objective is computing click-through rates, which are essentially probabilities that a certain user will click on a specific displayed ad. Since the act of clicking on an ad is an event with a binary outcome, we can formulate our problem as binary classification, where a value of $1$ implies that the user clicked the ad and a value of $0$ implies they didn't. Our models are trained to predict the event outcome by using the rest of the event information as dependant variables. 

\subsubsection{Implementation}
We implement our LR, FM and DeepFM baseline models as well as all embedding modules in the programming language Python~\cite{python}. Python is an object oriented, scripted and interpreted language, which is currently considered one of the premier tools for data scientists, both in education, as well as general research~\cite{Paffenroth2015PythonID}. In addition to its ease of use, one of Python's main advantages is its access to a large array of programming libraries. To implement our models and carry out our experiments, we primarily utilize TensorFlow~\cite{tensorflow}, an end-to-end open source machine learning library. To ease the use of our work for solving other problems, we employ a modular approach, where we use a single configurable function to construct all of our models. The function has the following parameters:
\begin{itemize}
    \item \textit{num\_feats} - how many features does the dataset have. Features are expected to be 32-bit integers.
    \item \textit{num\_bins} - the size of the hashing domain for the entire dataset. This setting implies that after hashing, each observed value in our dataset will range between $0$ and $\textrm{\textit{num\_bins} - 1}$.
    \item \textit{num\_factors} - the size of the FM model feature interaction vectors. If this parameter is not set, or the size is set to $0$, the model becomes LR.
    \item \textit{num\_hidden\_layers} - how many layers are in the neural network part of the DeepFM model. If the value is set to $0$, the model becomes LR/FM.
    \item \textit{hidden\_layer\_size} - number of neurons in a single layer of the neural network part of the DeepFM model. If the value is set to $0$, the model becomes LR/FM.
    \item \textit{linear\_modules} - the embedding modules that are applied to the linear embeddings.
    \item \textit{interaction\_modules} - the embedding modules that are applied to the interaction embeddings.
    \item \textit{both\_modules} - the embedding modules that are applied to both linear and interaction embeddings.
    \item \textit{optimizer} - the optimizer used to update model parameters during training.
    \item \textit{loss} - the loss used to guide optimization.
    \item \textit{additional\_metrics} - additional informative metrics to display during model training. 
\end{itemize}
%The baseline model, all proposed embedding modules, as well as evaluation code are open-source and available at \url{https://github.com/Kahno/feature_embedding} under the BSD-3-Clause license.

\subsubsection{The dataset}
Our experiments are performed on a publicly available dataset provided by Criteo \cite{criteo,criteodata}. The dataset represents a portion of their traffic over a period of 7 days. It contains 45 million examples of served display ads with 26 categorical features and the target variable; whether a click occurred. Categorical data is anonymized and presented in the form of hashed values. Positive and negative samples have both been subsampled at different rates beforehand, resulting in a final ratio of 26\% click and 74\% non-click events. Since the dataset simulates a scenario of real-time ad space bidding, we are not allowed to perform any kind of sample shuffling. Doing so would imply that the model predicts present data, but is trained with samples from the future. To enforce the time and order sensitive nature of our dataset, we select the data samples from the first 70\% as our training set, and use the remaining 30\% as our test set.

% Removed sections:
%\subsubsection{Activation functions}
%\subsubsection{Weight initialization procedures}
%\subsubsection{Combinations of activations and initalizations}

\subsubsection{Parameter and hyperparameter optimization}
All our models were trained with the LazyAdam optimizer, a variant of the popular Adam~\cite{kingma2014adam} optimizer, which is better suited for handling sparse updates. We used grid search to explore different hyperparameter configurations for each embedding module as well as the baseline models. For FM and DeepFM models, we selected a fixed latent vector size of 6. We performed separate grid searches for the optimal learning rate of the LR and FM models. For the baseline DeepFM model, we performed a grid search of the optimal values of hidden layer size, number of hidden layers and learning rate. The three obtained optimal models served as baselines. We performed a series of experiments where we applied each individual module to our model baselines and performed a grid search of the respective module's hyperparameters. We explored the following hyperparameters:
\begin{itemize}
    \item Embedding scaling: scaling ratio (scaling values 0.1, 0.2, 0.3, 0.4, 0.5, 0.75, 1, 2, 3), number of hidden layers (1, 2, 3), and layer activation (ReLU, Swish, Tanh).
    \item FM embedding: size of new latent vectors (2, 3, 4, 5, 6, 7, 8, 9, 10).
    \item Embedding encoding: squeeze ratio (1.5, 2, 3, 4, 5, 6), number of hidden layers (1, 2, 3), and layer activation (ReLU, Swish, Tanh).
    \item NN embedding: number of hidden layers (1, 2, 3) and layer activation (ReLU, Swish, Tanh).
\end{itemize}
In addition to the module's hyperparameters, we also explored a new optimal learning rate for each module-enhanced model. All our experiments feature a batch size of $10000$.

\subsubsection{Model evaluation}
Due to the class imbalance present in our dataset, using measures such as precision is undesirable. A naive model that always predicts the majority class would achieve $74\%$ precision on our dataset, but would be entirely useless at the task of click prediction. Instead of precision, our primary performance metric is therefore Relative Information Gain~\cite{graepel2010web-scale}. We first define the empirical cross entropy or log-score as follows:
\begin{equation*}
    CE = \frac{1}{N} \sum^N_{i=1} \Big[ y_i \log p_i + (1-y_i)\log(1-p_i) \Big],
\end{equation*}
where $y_i$ equals the label of the $i$-th sample, $p_i$ equals the predicted probability of the $i$-th sample and $N$ equals the number of test samples. Given the empirical CTR of the data $p = \sum^N_{i=1} y_i / N$, we define the information gain as $IG = CE + H(p)$, where $H$ is the entropy defined by:
\begin{equation*}
    H(p) = -(p \log p + (1-p)\log(1-p)).
\end{equation*}
We define relative information gain (RIG) as the ratio $RIG = IG/H(p)$.

\section{results}

\subsection{Logistic regression}
We present the following LR-based models:
\begin{itemize}
    \item LR; the baseline model,
    \item LR+Scale; LR model augmented with the embedding scaling module,
    \item LR+Encode; LR model augmented with the embedding encoding module,
    \item LR+Weight; LR model augmented with the embedding reweighting module.
\end{itemize}
Augmented models LR+Scale and LR+Encode use the original linear embeddings as input to generate new embeddings. LR+Weight similarly takes the original linear embeddings as input and generates a vector that reweights each embedding value. Our grid search finds the following optimal hyperparameter values for each augmented model:
\begin{itemize}
    \item LR+Scale has scaling factor $1.5$, two hidden layers, ReLU activation and learning rate $0.005$,
    \item LR+Encode has dimension scaling value $1.5$, one hidden layer, Swish activation and learning rate $0.009$.
    \item LR+Weight has a learning rate of $0.01$
\end{itemize}
The baseline LR model has a learning rate of $0.003$. The results of each model with the described configurations can be seen in Table~\ref{tab:lr_results}.

\begin{table}[ht]
\centering
\scalebox{1}{
\begin{tabular}{l r c l r}
 \toprule
 Model & RIG [\%] & Log loss [\%] & Training time\\
 \midrule
 LR & $16.06 \pm 0.00$ & $47.96 \pm 0.00$ & 4min 10s\\
 \midrule
 LR+Scale & $16.58 \pm 0.05$ & $47.67 \pm 0.03$ & 5min 37s\\
 LR+Encode & $16.40 \pm 0.07$ & $47.77 \pm 0.04$ & 5min 8s\\
 LR+Weight & $16.76 \pm 0.01$ & $47.56 \pm 0.01$ & 4min 46s\\
 \bottomrule
\end{tabular}
}
\caption{\textbf{Logistic regression results.} Relative information gain and log loss are expressed as percentage values.}
\label{tab:lr_results}
\end{table}

All proposed augmented models provide a performance increase over the LR baseline. Since each augmented model features a different type of embedding module, we are able to observe their contributions. The performance contribution of each embedding module can be seen in Table~\ref{tab:lr_contribution}. 

\begin{table}[ht]
\centering
\scalebox{1}{
\begin{tabular}{l r}
 \toprule
 Module & Performance increase\\
 \midrule
 Embedding scaling & $0.52 \pm 0.05$\\
 Embedding encoding & $0.34 \pm 0.07$\\
 Embedding reweighting & $0.70 \pm 0.01$\\
 \bottomrule
\end{tabular}
}
\caption{\textbf{Module contributions for the LR experiment.} Performance increase is measured in percentage points of the RIG metric.}
\label{tab:lr_contribution}
\end{table}

The best performing module is the embedding reweighting module. It provides the most significant improvement in terms of RIG, while not drastically increasing prediction time. Notably, such offline performance improvements could translate to a significant increase in online\footnote{Offline implies a local training dataset, while online implies live production data.} predictive performance~\cite{DBLP:journals/corr/ChengKHSCAACCIA16}. In addition to the embedding reweighting module, both the embedding scaling and embedding encoding modules also provide a substantial increase in predictive performance.

\subsection{Factorization machines}
We present the following FM-based models:
\begin{itemize}
    \item FM; the baseline model,
    \item FM+Scale; FM model augmented with the embedding scaling module,
    \item FM+FM; FM model augmented with the FM embedding module,
    \item FM+Encode; FM model augmented with the embedding encoding module,
    \item FM+NN; FM model augmented with the NN embedding module,
    \item FM+Weight; FM model augmented with the embedding reweighting module.
\end{itemize}

\begin{table}[ht]
\centering
\scalebox{1}{
\begin{tabular}{l r r r r}
 \toprule
 Model & RIG [\%] & Log loss [\%] & Training time\\
 \midrule
 FM & $17.03 \pm 0.02$ & $47.41 \pm 0.00$ & 9min 13s\\
 \midrule
 FM+Scale & $17.59 \pm 0.00$ & $47.09 \pm 0.00$ & 9min 43s\\
 FM+FM & $17.11 \pm 0.02$ & $47.36 \pm 0.01$ & 30min 46s\\
 FM+Encode & $17.47 \pm 0.02$ & $47.16 \pm 0.01$ & 13min 55s\\
 FM+NN & $17.33 \pm 0.13$ & $47.24 \pm 0.07$ & 15min 11s\\
 FM+Weight & $17.71 \pm 0.01$ & $47.02 \pm 0.00$ & 10min 22s\\
 \bottomrule
\end{tabular}
}
\caption{\textbf{Factorization machine results.} Relative information gain and log loss are expressed as percentage values.}
\label{tab:fm_results}
\end{table}

Augmented models FM+Scale, FM+Encode and FM+Weight apply the effects of their respective embedding modules to both linear and interaction embeddings. Afterwards, all embeddings are used to compute the final prediction. The FM+FM model only transforms the interaction embeddings. The FM+NN model uses both linear and interaction embeddings as a single input to compute new versions of both embeddings. Our grid search finds the following optimal hyperparameter values for each augmented model:
\begin{itemize}
    \item FM+Scale has scaling factor $0.3$, one hidden layer, Swish activation and learning rate $0.004$,
    \item FM+FM has $9$ factors in the FM used for the embedding and learning rate $0.003$,
    \item FM+Encode has dimension scaling value $3$, three hidden layers, Swish activation and learning rate $0.007$,
    \item FM+NN has two hidden layers, ReLU activation and learning rate 0.006,
    \item FM+Weight has a learning rate of $0.004$.
\end{itemize}
The baseline FM model has a learning rate of $0.001$. The results of each model with the described configurations can be seen in Table~\ref{tab:fm_results}.

\begin{table}[ht]
\centering
\scalebox{1}{
\begin{tabular}{l r}
 \toprule
 Module & Performance increase\\
 \midrule
 Embedding scaling & $0.57 \pm 0.02$\\
 FM embedding & $0.09 \pm 0.03$\\
 Embedding encoding & $0.44 \pm 0.03$\\
 NN embedding & $0.30 \pm 0.13$\\
 Embedding reweighting & $0.68 \pm 0.02$\\
 \bottomrule
\end{tabular}
}
\caption{\textbf{Module contributions for the FM experiment.} Performance increase is measured in percentage points of the RIG metric.}
\label{tab:fm_contribution}
\end{table}

All proposed augmented models provide a performance increase over the FM baseline. Similarly to our LR experiments, each FM augmented model features a different type of embedding module, so we again observe their individual contributions. The performance contributions of each embedding module can be seen in Table~\ref{tab:fm_contribution}.

The best performing module is again the embedding reweighting module. It provides a similar RIG increase as noted in the LR experiments, with model training time again not increasing drastically. The embedding scaling and embedding encoding modules also provide a similar performance increase as noted in the LR experiments. The newly presented NN embedding module is also reasonably successful. The FM embedding module is the least successful, but still provides a small lift. The training time is notably increased in this case.

\subsection{Deep factorization machines}
We present the following DeepFM-based models:

\begin{table}[ht]
\centering
\scalebox{1}{
\begin{tabular}{l r r r r}
 \toprule
 Model & RIG [\%] & Log loss [\%] & Training time\\
 \midrule
 DeepFM & $17.73 \pm 0.01$ & $47.01 \pm 0.01$ & 18min 24s\\
 \midrule
 DeepFM+Scale & $17.83 \pm 0.01$ & $46.95 \pm 0.01$ & 23min 42s\\
 DeepFM+FM & $17.62 \pm 0.01$ & $47.07 \pm 0.01$ & 52min 27s\\
 DeepFM+Encode & $17.70 \pm 0.02$ & $47.03 \pm 0.01$ & 22min 33s\\
 DeepFM+NN & $17.78 \pm 0.02$ & $46.98 \pm 0.01$ & 23min 26s\\
 DeepFM+Weight & $17.83 \pm 0.01$ & $46.95 \pm 0.01$ & 20min 10s\\
 \bottomrule
\end{tabular}
}
\caption{\textbf{DeepFM results.} Relative information gain and log loss are expressed as percentage values.}
\label{tab:deepfm_results}
\end{table}

\begin{itemize}
    \item DeepFM; the baseline model,
    \item DeepFM+Scale; DeepFM model augmented with the embedding scaling module,
    \item DeepFM+FM; DeepFM model augmented with the FM embedding module,
    \item DeepFM+Encode; DeepFM model augmented with the embedding encoding module,
    \item DeepFM+NN; DeepFM model augmented with the NN embedding module,
    \item DeepFM+Weight; DeepFM model augmented with the embedding reweighting module.
\end{itemize}
The module-enhanced models match their FM counterparts, with the addition of the deep neural network component. Our grid search finds the following optimal hyperparameter values for each augmented model:
\begin{itemize}
    \item DeepFM+Scale has scaling factor $3$, one hidden layer, ReLU activation and learing rate $0.001$,
    \item DeepFM+FM has $8$ factors in the FM used for the embedding and learning rate $0.002$,
    \item DeepFM+Encode has dimension scaling value $1.5$, one hidden layer, ReLU activation and learning rate $0.001$,
    \item DeepFM+NN has one hidden layer, ReLU activation and learning rate $0.001$,
    \item DeepFM+Weight has a learning rate of $0.004$.
\end{itemize}
The DeepFM+Weight, DeepFM+Scale and DeepFM+NN models are able to provide a performance increase over the DeepFM baseline. The DeepFM+Encode as well as DeepFM+FM decrease model performance. Exactly as above, each DeepFM augmented model features a different type of embedding module, so we can observe their individual contributions. The performance contributions of each embedding module can be seen in Table~\ref{tab:deepfm_contribution}.

\begin{table}[ht]
\centering
\scalebox{1}{
\begin{tabular}{l r}
 \toprule
 Module & Performance increase\\
 \midrule
 Embedding scaling & $0.10 \pm 0.01$\\
 FM embedding & $-0.11 \pm 0.01$\\
 Embedding encoding & $-0.03 \pm 0.02$\\
 NN embedding & $0.05 \pm 0.02$\\
 Embedding reweighting & $0.11 \pm 0.01$\\
 \bottomrule
\end{tabular}
}
\caption{\textbf{Module contributions for the DeepFM experiment.} Performance increase is measured in percentage points of the RIG metric.}
\label{tab:deepfm_contribution}
\end{table}

The best performing module is once again the embedding reweighting module. It provides a substantially lower performance increase than in previous experiments, which is likely due to the already impressive performance of the baseline model. DeepFMs are currently considered among the state-of-the-art models for CTR prediction. 

\section{discussion} 
In this manuscript, we investigate different feature embedding approaches when dealing with high-dimensional categorical data. We specifically aim to improve predictive performance of logistic regression, factorization machines and deep factorization machines for the task of CTR prediction. We propose five different embedding modules, apply them to the relevant baseline models and evaluate their predictive performance and training time on a popular public CTR dataset. Our experiments suggest that multiple proposed embedding modules provide a significant performance improvement over the model baselines. 

For both logistic regression and factorization machines, the most successful modules are the embedding re\-we\-ight\-ing, embedding scaling and embedding encoding modules, which add a significant boost to the baseline models' RIG performance. Particular to the factorization machine model we evaluate two additional embedding modules: FM and NN embedding. The performance of NN embedding is slightly lower compared to embedding encoding, while FM embedding only provides a minor RIG lift.

In the DeepFM experiment, we observe positive performance of NN embedding, scaling and reweighting, with the latter again giving us the best results. With DeepFM, embedding encoding and FM embedding are unable to improve the baseline performance. Applying the FM embedding module to the baseline models also results in significantly longer training time, while other embedding modules do not produce a noticeable increase.

Our experiments present the embedding reweighting module as the clear winner in terms of performance boosting. Since the module essentially learns to assign a weight to each feature, further research could be made to investigate and build a framework that monitors the reweighting layer of the model and identifies features with higher and lower influence. This information could then be utilized to perform feature selection in a production setting.

Finally, there remain two major areas of further research: investigating how different combinations of proposed modules affect performance when combined into a single model and applying the proposed feature embedding findings to other machine learning areas. Our work investigates the feature embedding effects for the task of click-through rate prediction, however the modules themselves can easily be applied to other types of models that also learn from tabular data. Furthermore, the enhanced models do not have to be limited to binary classification, which further broadens the field of potential applications. 

\section*{conflict of interest}

The authors declare that the research was conducted in the absence of any commercial or financial relationships that could be construed as a potential conflict of interest.

\section*{code availability}

The baseline model, all proposed embedding modules, as well as evaluation code are available at \url{https://github.com/Kahno/feature_embedding}. This is an open source project, licensed under the BSD-3-Clause license.

\bibliographystyle{vancouver}
\bibliography{click-prediction}

\end{document}